\pdfoutput=1

\documentclass[11pt]{article}

\usepackage[]{ACL2023}

\usepackage{times}
\usepackage{latexsym}
\usepackage{booktabs}
\usepackage{multirow}
\usepackage{amsmath,amssymb,amsfonts}
\usepackage{threeparttable}
\usepackage{algorithmic}
\usepackage{graphicx}
\usepackage{textcomp}
\usepackage{utfsym}
\usepackage{xcolor}
\usepackage[most]{tcolorbox}
\usepackage[T1]{fon tenc}

\usepackage[utf8]{inputenc}

\usepackage{microtype}

\usepackage{inconsolata}

\title{InstructERC: Reforming Emotion Recognition in Conversation with Multi-task Retrieval-Augmented Large Language Models}

\author{
    Shanglin Lei\textsuperscript{1}\thanks{Equal contribution},
    Guanting Dong\textsuperscript{2}$^{*}$,
    Xiaoping Wang\textsuperscript{1}\thanks{Corresponding author},
    Keheng Wang\textsuperscript{3}
    Runqi Qiao\textsuperscript{2},
    Sirui Wang\textsuperscript{4} \\
    \textsuperscript{1}School of Artificial Intelligence and Automation, HUST, Wuhan, China \\
    \textsuperscript{2}School of Artificial Intelligence, BUPT, Beijing, China \\
    \textsuperscript{3}Sino-French Engineer School, Beihang University, Beijing, China \\
    \textsuperscript{4}Meituan Inc., Beijing, China \\
}

\begin{document}
\maketitle

\begin{abstract}
The field of emotion recognition of conversation (ERC) has been focusing on separating sentence feature encoding and context modeling, lacking exploration in generative paradigms based on unified designs.
In this study, we propose a novel approach,
 \textbf{InstructERC}, to reformulate the ERC task from a discriminative framework to a generative framework based on Large Language Models (LLMs).
 InstructERC makes three significant contributions: 
(1) it introduces a simple yet effective retrieval template module, which helps the model explicitly integrate multi-granularity dialogue supervision information.
(2) We introduce two additional emotion alignment tasks, namely speaker identification and emotion prediction tasks, to implicitly model the dialogue role relationships and future emotional tendencies in conversations.
(3) Pioneeringly, we unify emotion labels across benchmarks through the feeling wheel to fit real application scenarios. InstructERC still perform impressively on this unified dataset.
Our LLM-based plugin framework significantly outperforms all previous models and achieves comprehensive SOTA on three commonly used ERC datasets. 
Extensive analysis of parameter-efficient and data-scaling experiments provides empirical guidance for applying it in practical scenarios. You can find the offical realization in the Github link: \url{https://github.com/LIN-SHANG/InstructERC}

\end{abstract}

\section{Introduction}
``The question is not whether intelligent machines can have emotions, but whether machines without emotions can achieve intelligence'', as mentioned in ``Society of Mind'' \cite{minsky1988society}.
Empowering machines with the ability to understand emotions in various scenarios has always been the unwavering direction of researchers. 

In contrast to conventional binary sentiment analysis tasks \cite{pontiki2016semeval} , which only rely on text with explicit attitude tendencies, the emotion recognition in conversation (ERC) task aims to identify more fine-grained emotional tendencies in each sentence of a conversation. 
Specifically, for a given complete dialogue sequence input and a set of emotional labels, the model is required to accurately assign an emotional label to each sentence.
Intuitively, the recognition of emotional tendencies in the target sentence is heavily influenced by its historical utterances \cite{yingjian2023emotionic}, and there is significant variation in how different speakers perceive and express emotions \cite{shen2021directed}.
Therefore, it is imperative to meticulously model the speakers and dialogue context. 

\begin{figure*}[t]
\centering
\includegraphics[width=1.8\columnwidth]{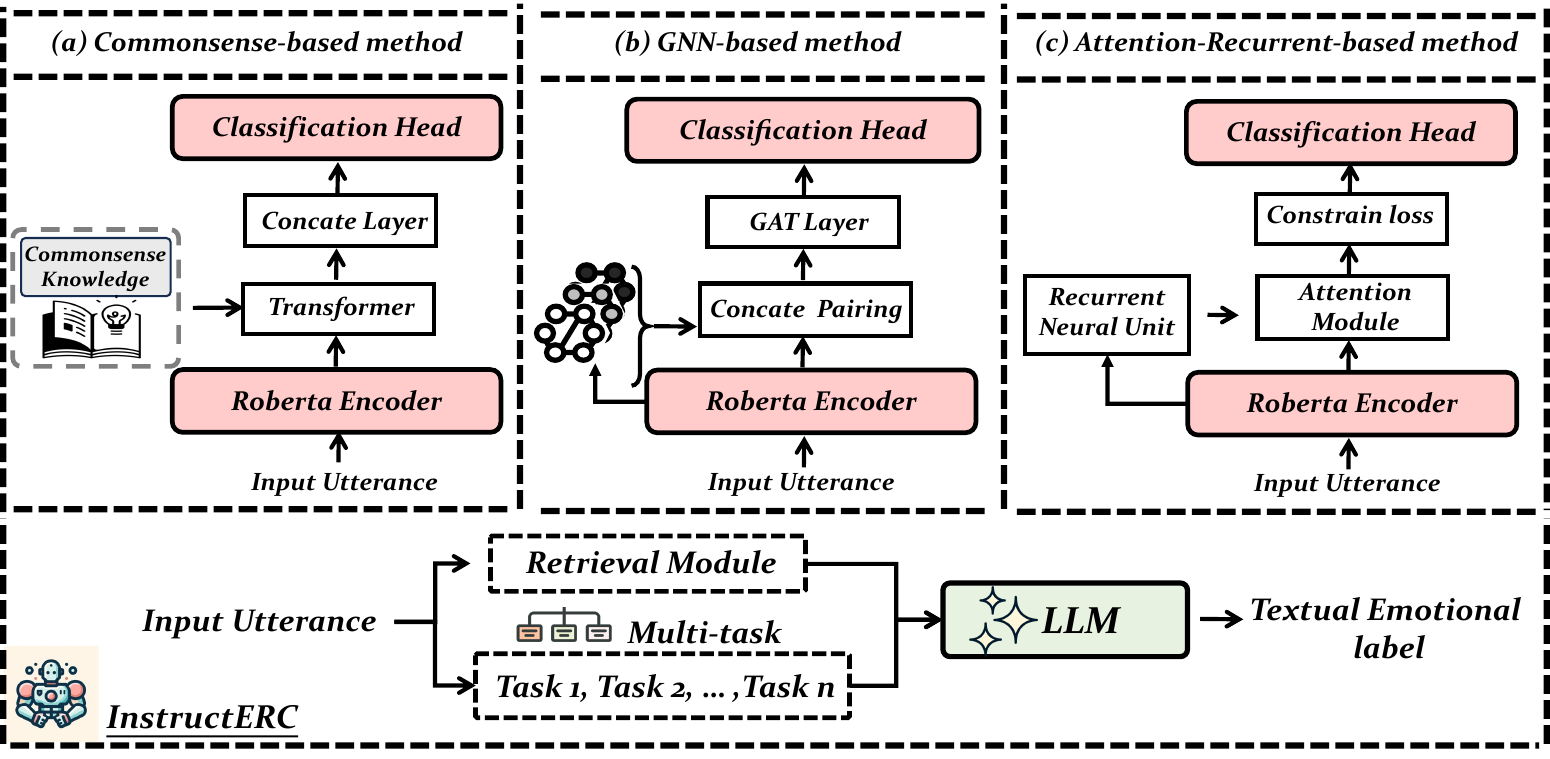}
\caption{The illustration of different paradigms for ERC}
\label{fig1}
\end{figure*}

Figure \ref{fig1} illustrates that previous work based on Roberta \cite{liu2019roberta} in ERC can be roughly divided into three categories: 
(1) \textbf{Transformer-based methods} \cite{li2020multi,song2022supervised,liu2023hierarchical,chudasama2022m2fnet} attempt to establish long-range emotional correlations in conversational scenarios by directly adopting or modifying the original transformer block. 
(2) \textbf{Recurrent-based methods} 
 \cite{hu2023supervised,lei2023watch,majumder2019dialoguernn,hazarika2018icon,poria2017context} utilize various forms of RNNs, like LSTM and GRU, to model individual emotional states and global emotional impacts separately.
(3) \textbf{GNN-based methods} \cite{ghosal2019dialoguegcn,ishiwatari2020relation,shen2021directed,li2023graphcfc}  typically use nodes and edges to model characters and dialogue relationships in conversations. 
Above approaches have their strengths in modeling dialogue at the sentence level, but they still generally adhere to the paradigm of fine-tuning sentence features and separately modeling dialogue context. However, in realistic scenarios, end-to-end model designs are often more practical. \footnote{The discussion between discriminant model and InstructERC can refer to \ref{appendix:Explanation_end_to_end}}.

Fortunately, the recent successful application \cite{openai2023gpt4,qwen2} and emergence capabilities \cite{zhao2023survey} of pre-trained large language models (LLMs) have demonstrated remarkable performance in natural language reasoning tasks. By using a generative architecture, LLMs unify the output and input of different tasks and have shown significant performance improvements in all NLP tasks. Despite their powerful capabilities, enabling these abilities for specific sub-tasks requires high-quality prompts \cite{wei2021finetuned,chung2022scaling} and designs to fill the reasoning gap. 
Therefore, how to use LLMs framework to reconstruct ERC while considering context modeling, speaker modeling, and capturing conversation relationships poses a significant challenge in pushing this framework towards a realistic ERC application.

In this work, we reformulate the ERC task using LLMs. Specifically, we design a simple but efficient retrieval template module, which consists of instruction, historical utterance, label statements, and demonstration retrieval to explicitly integrate multi-granularity dialogue supervision information during reasoning. In addition, we separately design two auxiliary tasks for the ERC task: speaker identification task and emotion prediction task. The speaker identification task assists LLMs in modeling dialogue role relationships by predicting the speaker of each sentence, while the emotion prediction task models future emotional tendencies in conversations. Furthermore, due to biases in data distribution and labeling across different ERC domains, it’s still challenging for discriminative ERC models to achieve multi-domain ERC capabilities, both in terms of engineering and performance. To dive deeper into this topic, we pioneeringly align labels for three benchmarks and conduct a series of unified dataset experiments. 
Looking ahead, we contend that IERC, as the first framework transitioning from single-domain to multi-domain ERC, offers us a glimpse into the prospective landscape of open-domain emotional artificial intelligence (Emotional AGI).


In conclusion, our work can be outlined as follows:
\begin{itemize}
\item To the best of our knowledge, we are the first to reformulate the ERC task as a retrieval based Seq2Seq paradigm with LLMs and present an effective instruction template which can adapt to different dialog scenarios.
\item We propose two novel emotional auxiliary tasks to implicitly model the dialogue role relationships and future emotional tendencies in conversations.
\item Our InstructERC significantly outperforms all previous models and achieves comprehensive SOTA on three commonly used ERC datasets. 
\item To advance towards multi-domain ERC scenario, we pioneeringly align labels for three benchmark to form the UIME ERC dataset, a series of unified dataset experimental results provides empirical guidance for application in practical scenarios.
\end{itemize}

\section{Methodology}
In this section, we present a comprehensive overview of the proposed InstructERC framework shown as Figure \ref{fig3}. Firstly, we provide a brief introduction to the task definition of ERC. Next, we discuss the framework of InstructERC, which consists of two major parts:
retrieval template module and emotional alignment tasks.
Finally, we introduce training and inference process of our framework. \footnote{Due to the space limitation, we have included the related works in Appendix \ref{appendix:related works}.}


\subsection{Problem Definition}\footnote{The difference of problem definition between two paradigms can be referred to in Appendix \ref{appendix:Explanation_end_to_end}.}
Assuming a dialogue text $U=[u_1,u_2,...u_n]$ of length $n$ is given, which includes $M$ speakers/parties $p_1, p_2,...,p_M$ ($M \geq 2$) in the dialogue, and each utterance $u_i$ spoken by the corresponding speaker $p_{K(u_i)}$. Function $K$ is employed to establish a mapping between each utterance and its corresponding speaker. $o$ is the number of emotional categories, which varies with the number of emotional types in different evaluation datasets.


\subsection{Retrieval Template Module}
\begin{figure}[t]
\centering
\includegraphics[width=1.0\columnwidth]{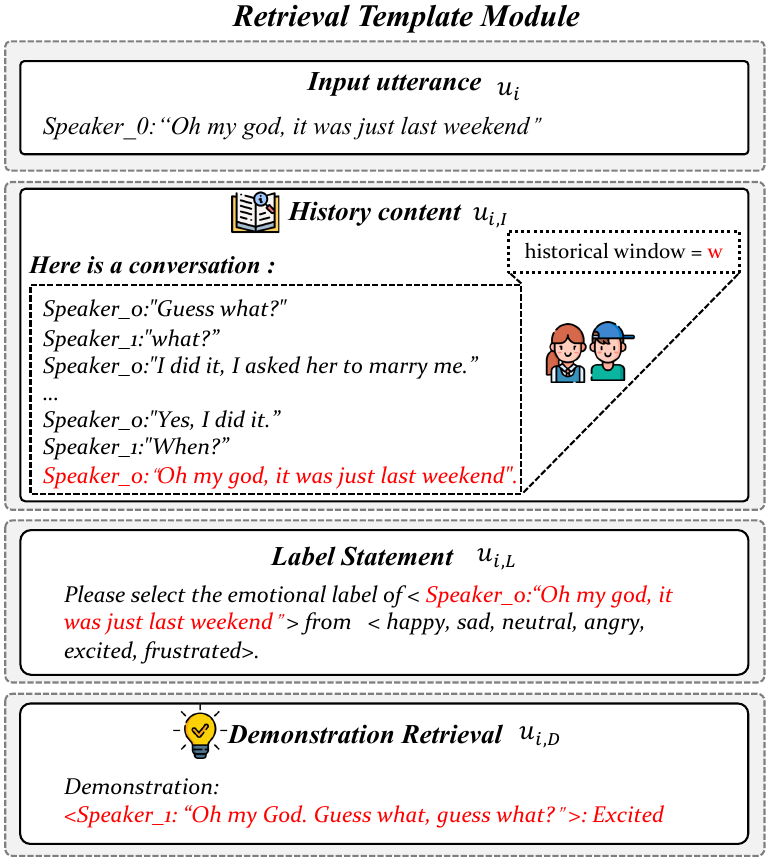} 
\caption{The Schematic of Retrieval Template Module.}
\vspace{-0.1cm}
\label{fig2}
\end{figure}

To better transfer and utilize the inference ability of pre-trained large language models, we reconstruct the ERC task to the seq2seq form and solve it through fine-tuning LLMs. Therefore, we construct a efficient retrieval template module to bridge the gap when applying LLMs to specific NLP subtasks. As shown in Figure \ref{fig2}, for ERC task, each input consists of four parts: instructions, historical content, label statement, and demonstration retrieval.

\textbf{Instruction.} The instructions serve to provide the model with a well-defined role, precise details of the ERC task, and a standardized format for the input dialogue text. For the primary ERC task, our instruction $u_{i,I}$ is shown in Figure \ref{fig2}. 

\textbf{Historical Content.} 
To model the context in realistic ERC scenarios,
We employ a hyperparameter, the historical window (denoted as $w$), to indicate the specific rounds (including current utterance) of historical dialogue along with the corresponding speaker information. For the emotion recognition of the target utterance $u_n$, its historical content $u_{i,H}$ is shown in Figure \ref{fig2}.

\textbf{Label Statement.} To confine the model's output within a finite range of labels 
and enable the model to focus on the current utterance being recognized, our label statement $u_{i,L}$ is shown in Figure \ref{fig2}. 

\textbf{Demonstration Retrieval.}
In order to further integrate emotional information to assist reasoning, we have developed a domain demonstration recall module based on semantic similarity. In detail, we construct a domain base $\mathcal{D}_{domain}$ from the training dataset that removes speaker identity information and balances the number of emotion labels, which ensures that the demonstrations is not influenced by the distribution of speakers or emotion labels in the dataset. For a given utterance $u_i$ to be identified, we retrieve the most relevant ERC example from $\mathcal{D}_{domain}$ as the demonstration. To perform the retrieval, we use a bidirectional encoder SBERT \cite{reimers2019sentence} to find the most semantically similar ERC example $d_{rvl}$. SBERT generates independent CLS embeddings for the target utterance $u_i$ and each element $d_j$ in $\mathcal{D}_{domain}$. After sorting all target-demonstration pairs by cosine similarity, we select the pair with the highest score as the most relevant element $d_{rvl}$. An abstract mathematical description of this process is as follows:
\begin{equation}
d_{rvl_i} = \mathop{\mathrm{argmax}}\limits_{d_j \in \mathcal{D}_{domain}} \mathrm {SBERT}(u_i , d_j) \label{eq:1}
\end{equation}
The textual input $u_{i,D}$ for the demonstration retrieval part is shown in Figure \ref{fig2}.
In summary, after constructing the Retrieval template, the simplified input $x_i$ for the main task is as follows:
\begin{equation}
x_i = [u_{i,I}; u_{i,H}; u_{i,L}; u_{i,D}] \label{eq:2}
\end{equation}
where [;] means the textual concatenation, $u_{i,I}$, $u_{i,H}$, $u_{i,L}$, and $u_{i,D}$ indicate Instructions, Historical content, Label statement, demonstration retrieval for a given utterance $u_i$.

\subsection{Emotional Alignment Tasks}
\begin{figure*}[t]
\centering
\includegraphics[width=2.0\columnwidth]{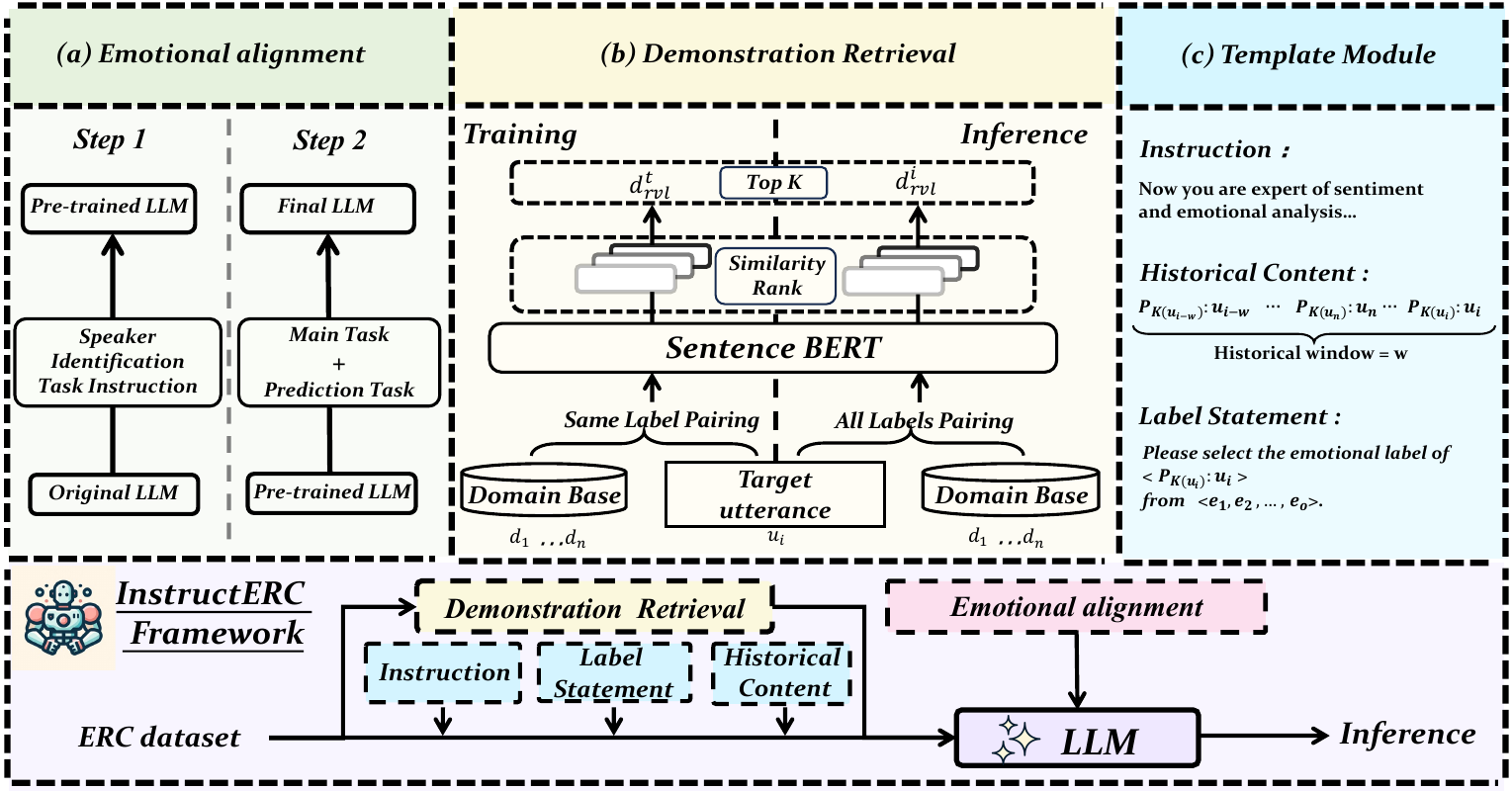} 
\vspace{-0.1cm}
\caption{The overview of InstructERC framework}
\label{fig3}
\vspace{-0.122cm}
\end{figure*}

To better capture the dialogue role relationships and future emotional tendencies in conversations, we have incorporated two auxiliary tasks, namely speaker identification and emotion impact prediction, which constitute the fine-grained subtasks of the InstructERC framework. The model is jointly trained with these auxiliary tasks to improve its overall performance, illustrated in Figure \ref{fig3}.

\textbf{Speaker Identification.}
Emotions are expressed differently among different speakers.
Previous models have used techniques such as speaker-based masked attention modules or multiple GRUs to capture the emotional expression features of different characters. 
This modeling of emotional expression in the task can also be transformed into a generative task using our InstructERC. 
To enable the LLM to capture the speaking styles of different individuals, beyond \cite{li2020multi}, the model is trained to identify the relevant speaker for a given utterance, without considering the historical context. For a given dataset, a predefined set of speaker labels is provided. Consistent with the main task, the Instruction text input $x^{p}_i$ for this task is constructed as follows:
\begin{quote}
\textit{``Now you are an expert of sentiment and emotional analysis. Please select the Speaker label of the utterance \textless Speaker:$ u_i$\textgreater\quad from \quad\textless$p_1$,...,$p_M$\textgreater''}
\end{quote}
The loss function for the Speaker Identification is as follows:
\begin{equation}
\mathcal{L}_p = \sum^{N}_i -\log P(\mu_i | x^{p}_i, \theta_p) \label{eq:3}
\end{equation}
Here, $\mu_i$ represents the token of the corresponding speaker label for the given speaker identification task input sample $x^p_i$.
Unless otherwise specified, $N$ stands for the total number of utterances in the dataset, while $\theta_{*}$ represents the parameters of the LLM in different periods.

\textbf{Emotion Impact Prediction.}
In the daily conversations, the intricate relationships between individuals can have a significant impact on the emotional states of subsequent dialog. 
Prior research has attempted to address this issue by constructing a dialogue relationship graph and utilizing a complex graph neural network to model the emotional impacts of these relationships. 
However, these methods are often associated with a highly intricate data preprocessing pipeline and are susceptible to overfitting on certain datasets. 
To address these issues, we propose a generative framework for the emotion impact prediction task, which implicitly captures the interplay between dialogues and emotional impacts.

Specifically, the input for emotion impact prediction consists of three parts: instruction, historical content, and label statement. First, the instruction part of this task is kept consistent with the instruction part of the main task. Then, since the task requires predicting the impact of previous historical utterances on the current utterance, unlike the main task, the historical content  $u^e_{i,H}$ with a window of ``w'' will not include the current utterance. Correspondingly, to stay aligned with the original design intention of the task, the label statement of this task is modified as follows:
\begin{quote}
\textit{``Based on the above historical utterances, the next utterance is spoken by \textless $P_{K(u_{i})}$ \textgreater, please predict the emotion states of \textless $P_{K(u_{i})}$ \textgreater  from \textless $e_1$, $e_2$, ..., $e_o$ \textgreater:''}
\end{quote}
Hence, the overall input for emotion impact prediction is:
\begin{equation}
x^e_i = [u_{i,I};u^e_{i,H},u^e_{i,L}] \label{eq:4}
\end{equation}
The loss calculation for the emotion impact prediction task is as follows:
\begin{equation}
\mathcal{L}_e = \sum^{N}_i -\log P(\epsilon_i | x^{e}_i, \theta_e) \label{eq:5}
\end{equation}
Here, $\epsilon_i$ represents the emotional label token of the text label $e_i$ corresponding to the formatted input utterance $x_i$. 
\subsection{Overview of InstructERC}

To sum up the instruction based generative framework for ERC, given an input utterance $x_i$ after concatenating the retrieval template $d_{rvl}$ and a LLM, the model returns the logits $g_i$ and the generated text $y_i$ for the entire sentence, including both input and output tokens. This is represented by the following equation:
\begin{equation}
y_i, \mathbf{g_i} = {\rm LLM}(x_i,\theta_{all}) \label{eq:6} 
\end{equation}

Here, $\theta$ is the same as mentioned. The LLM predicts the conditional probability $p(\gamma_i| x_i, \theta)$ of generating each token $\gamma_i$ of the generated text $y_i$ until the end symbol \textless eos\textgreater  is outputted. 
As for logits $\mathbf{g_i} \in {{\mathbf{R}}^{L \times V}}$,
where $L$ and $V$ denote the length of the entire sentence and the size of the vocabulary used by the LLM, respectively.

In accordance with the original training method of LLMs, we adopt the next token prediction loss to measure the model's output error. Therefore, the loss calculation of the main task, denoted as $\mathcal{L}_{main}$, is defined as follows:

\begin{equation}
 \mathcal{L}_{main} = \sum^{N}_i -\log P(\epsilon_i| x_i, \theta_{all}) \label{eq:7} 
\end{equation}


\textbf{Training and Inference.} 
During training and inference, our retrieval process, emotional alignment tasks and main tasks in InstructERC can be divided into two stages:

In the first stage of joint training, the characteristics of the speaker intuitively form the basis of emotional expression. Therefore, we use the speaker identification task for LLM pre-training to fine-tune speaker characteristics, which aims to preheat parameters for subsequent ERC tasks. 

In the second stage, we fine-tune LLM using both the ERC main task and the emotion influence prediction task to improve overall performance. The training loss at this stage is $\mathcal{L}_{main}+\alpha * \mathcal{L}_e$,  where $\alpha$ is a hyperparameter used to adjust the weight of the emotion influence prediction task loss in the second overall joint training loss.

The difference of demonstration retrieval on training and inference stage is shown in figure \ref{fig3}, we limit the retrieved examples to those with the same emotion label as the current recognized speech, namely same label pairing ,in order to provide more diverse emotional understanding while avoiding excessive noise during training. During inference, there are no restrictions on the retrieved demonstrations due to the labels are unknown, namely All labels pairing. The retrieval results, simply referred as $d_{rvl}$, are specialized as $d^{t}_{rvl}$  and $d^{i}_{rvl}$ in training and inference stage, respectively.

\section{Experiments and Results}

\subsection{Dataset}
We evaluate the efficacy of InstructERC on three standard benchmark datasets: IEMOCAP, MELD, and EmoryNLP. The specifics of the datasets are outlined in Table \ref{tab:datasets_statics}. The details of dataset can be refer to Appendix \ref{appendix:datasets}.

\subsection{Baselines}
Align with the related works, we select several only textual modality baselines to compare with our InstructERC. \textbf{1) Transformer-based}: SPCL+CL\cite{song2022supervised} and MPLP \cite{zhang2023mimicking} , \textbf{2) Recurrent-based}: EmotionIC\cite{yingjian2023emotionic} and SACL-LSTM\cite{hu2023supervised}, \textbf{3) GNN-based}: 
DualGATs\cite{zhang-etal-2023-dualgats}
and Skier\cite{li2023skier}.
\textbf{4) LLM backbones}: ChatGLM-6B \& ChatGLM2-6B \cite{du2022glm} and LLaMA-7B \& LLaMA2-7B \cite{touvron2023llama}.
More details of baselines and implementations can be refered to Appendix \ref{appendix:baselines} and \ref{appendix:implementation}.


\begin{table}[htbp]
\caption{The main results on three benchmarks.}
\scalebox{0.75}{
\begin{tabular}{ccccc}
\toprule
 Dataset          & IEMOCAP                    & MELD                       & EmoryNLP  & Average                 \\
{Models}                   & {W-F1}       & {W-F1}       & {W-F1}   &    {W-F1} \\
\midrule 
 \multicolumn{5}{c}{Disciminant Models}       \\
\midrule 

{SPCL+CL}$^\dagger$ & 69.74                      & 66.35                      & 40.25  & {\color{orange}\textbf{\textit{58.78}}}     \\
{MPLP}$^*$    &  66.65             & 66.51                      & -        &  -               \\
{EmotionIC}$^\dagger$ & {\color{orange}\textbf{\textit{69.61}}}              & 66.40                      & 40.01   & {58.63}     \\
{SACL}$^*$ & 69.22                      &  66.45             & 39.65          & 58.44    \\
{DualGATs}$^*$     & 67.68	            & 66.90	                      & {\color{orange}\textbf{\textit{40.29}}}           & 58.29  \\
{Skier}$^\dagger$  & -                  & {\color{orange}\textbf{\textit{67.39}}}	                      & 40.07                 &  -     \\
\midrule
 \multicolumn{5}{c}{Zero-shot + InstructERC}       \\
\midrule
ChatGPT3.5$^\dagger$ & \textbf{\textit{53.38}} & \textbf{\textit{65.07}} & \textbf{\textit{37.00}} & \textbf{\textit{51.81}} \\
 ChatGLM$^\dagger$  & 38.6  & 38.8  & 19.6 & 32.33\\
ChatGLM2$^\dagger$ & 21.1  & 21.8  & \textbf{\textit{24.4}} & 22.43 \\
Llama$^\dagger$    & 0.753 & 9.12  & 5.31 & 5.06\\
Llama2$^\dagger$   & 2.774 & 16.28 & 8.36 & 9.46\\
\midrule
 \multicolumn{5}{c}{LoRA + Backbone}       \\
\midrule 

ChatGLM$^\dagger$                             & {17.98} & {40.54} & {25.71}  & 28.07\\
ChatGLM2$^\dagger$                            & {52.88} & {64.85}  & {37.69} & 51.80\\
Llama$^\dagger$                               & {55.81} & \textbf{\textit{66.15}}  & {37.98} & 53.21\\
Llama2$^\dagger$                              & \textbf{\textit{55.96}} & {65.84} & \textbf{\textit{38.21}} & \textbf{\textit{53.33}}\\
\midrule
 \multicolumn{5}{c}{LoRA + InstructERC}       \\
\midrule 
 
ChatGLM$^\dagger$                               & 36.04                   & 46.41                 & 30.86             & 37.77 \\
ChatGLM2$^\dagger$                               & 67.54                     & 65.58              & 39.09             & 57.40 \\
Llama$^\dagger$                                  & 64.17                     & 67.62              & 39.34             & 57.04 \\
Llama2$^\dagger$                                 & {\color{red} \textbf{71.39}}   & {\color{red}\textbf{69.15}}   & {\color{red}\textbf{41.37}}    & {\color{red}\textbf{60.64}} \\
\bottomrule
\end{tabular}
}
\begin{tablenotes}

\item[a] \tiny{NOTE: The best results of other baselines are in gold font, while SOTA results across all models are emphasized in red font. * indicate results sourced from the model's paper, and a ($\dagger$) denotes results from reproductions conducted by the authors. }
\end{tablenotes}
\vspace{-0.2cm}
\label{tab2}
\end{table}

\subsection{Main Results}
Table \ref{tab2} illustrates the results of comparing our InstructERC model with other models and backbones from different perspectives. Based on this, We make the following observations: 

(1) Our methods achieves significant improvements over the SOTA of discriminative models on all benchmarks. Specifically, we outperform EmotionIC, Skier, and DuaGATs by 1.73\%, 1.76\%, and 1.08\% on IEMOCAP, MELD and EmoryNLP respectively. Notably, we completely outperformed commonsense knowledge models (Skier) on two benchmarks without any external knowledge, demonstrating the extreme utilization of our method for textual data.

(2) To gain an insight into LLM models under different supervision scenarios for ERC task, we conduct experiments on Zero-shot + InstructERC and LoRA + InstructERC settings. It can be observed that even with carefully designed primary task instructions, LLMs still struggle in zero-shot scenarios, which further confirms the existence of a significant reasoning gap in their application to ERC sub-task.
Furthermore, by utilizing the LoRA + InstructERC, the performance of the four LLMs has significantly improved, especially on the IEMOCAP dataset. This fully demonstrates the effectiveness and generalization ability of our InstructERC framework, which greatly enhances the emotion recognition capability of LLM in long texts.

(3) InstructionERC is a plug-and-play method that can be adapted to multiple generative frameworks, such as prefix decoder or causal decoder. Although ChatGPT has a relevant competitive good performance on short length conversation scenrios(e.g. Meld,EmoryNLP), as can be seen, our results are far superior to the level of ChatGPT. Our unified alignment task and demonstration construction strategy are not tailored to any specific dataset design, highlighting the strong transferability and generalization capability of our approach.

\begin{figure*}
    \centering
    \includegraphics[width=2.0\columnwidth]{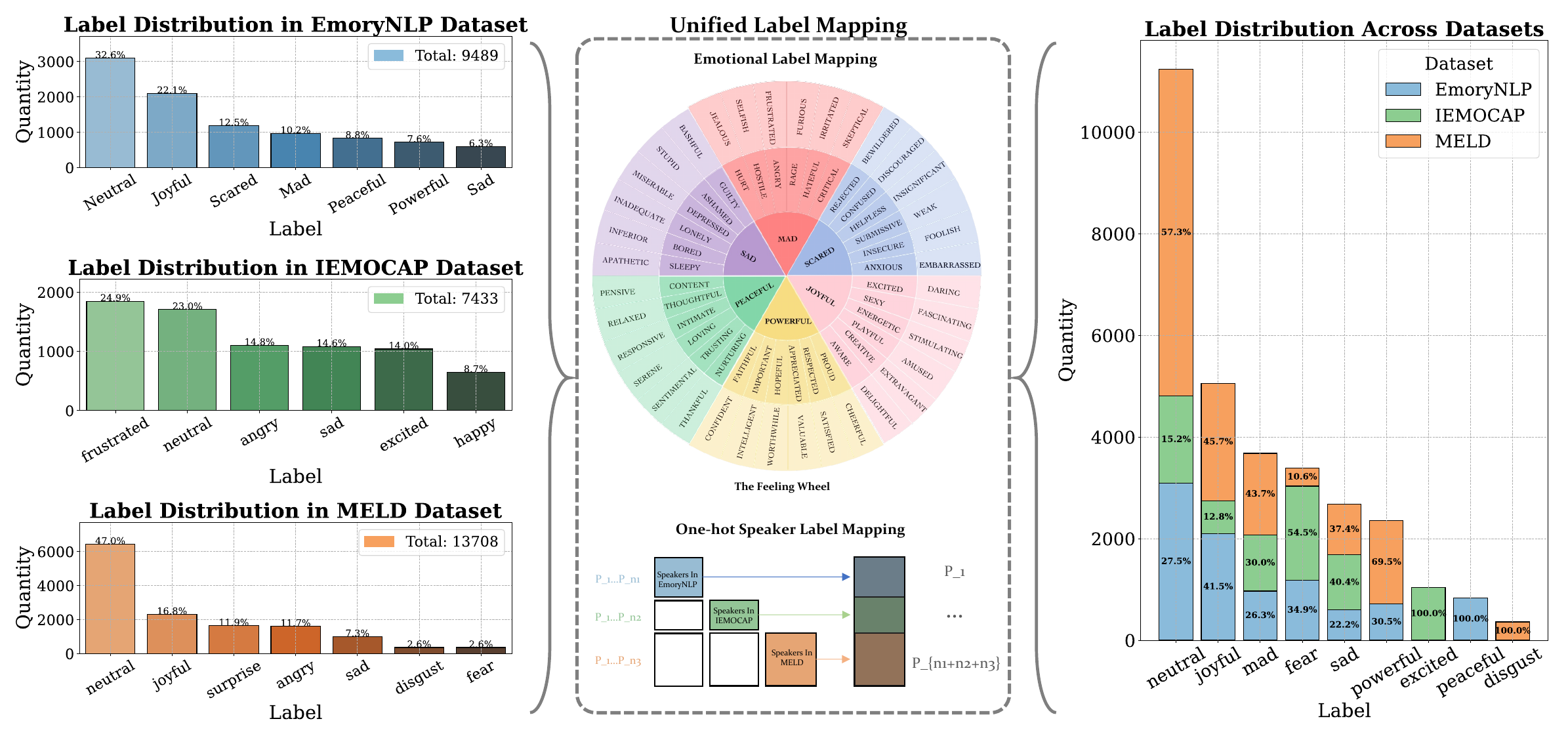}
    \caption{Unified Label Mapping Across three Open-source Benchmarks. The Feeling Wheel is proposed by \cite{willcox1982feeling}}
    \label{fig:Unified_mapping}
\end{figure*}

\subsection{Ablution study}

We conduct an ablation study to investigate the characteristics of the main components in InstructERC. Table \ref{tab3} shows the ablation results, and “w/o" denotes the model performance without a specific module. We have following observations: 

(1) The performance of InstructERC drops when removing any one component, which suggests that every part of the design is necessary. 

(2) Removing any one Emotional alignment task results in great performance degradation. This is consistent with our conjecture since speaker identification and emotion impact prediction provide relatively orthogonal semantic information from two perspectives. \footnote{We also explore the impact of $\alpha$ on the performance of InstructERC, refer to Appendix E.2}

(3) Taking away the domain retrieval module resulted in a steady decline on all three datasets, demonstrating the important role of domain information in dialogue modeling. 

4) Removing joint alignment task tasks causes obvious performance degradation compared with removing one of them, which indicates that jointly pre-training objectives have a mutually reinforcing effect. \footnote{We also explore the optimal conversational turns in modeling context in ERC, please refer to the``The historical window exploration study'' section in Appendix \ref{appendix:historical}.}

(5) Replacing LoRA with full-parameter fine-tuning results in a significant drop in performance, which indicates that the parameter-efficient approach is effective in preventing overfitting of LLMs on the ERC task. For detailed analysis, please refer to the ``All Parameters vs Parameter Efficiency'' section in Appendix \ref{appendix:parameters} . The further data scaling analysis of single dataset 
can be refer to Appendix \ref{appendix:single-scaling}.

\begin{table}[]
\caption{The ablation results of Llama2 on three benchmarks.}
\scalebox{0.8}{
\begin{tabular}{cccc}
\toprule
 Dataset          & IEMOCAP                    & MELD                       & EmoryNLP        \\ 
{Models}                   & {W-F1}       & {W-F1}       & {W-F1} \\ 
\midrule
 \multicolumn{4}{c}{LoRA + InstructERC}       \\
\midrule 
Llama2             & \textbf{71.39}$_{\pm 0.10}$ & \textbf{69.15}$_{\pm 0.08}$ & \textbf{41.37}$_{\pm 0.11}$  \\
w/o $\mathcal{L}_e$ & 70.50$^{**}$$_{\pm 0.12}$ & 68.97$^{*}$$_{\pm 0.10}$ & 40.78$^{*}$$_{\pm 0.10}$ \\
w/o $\mathcal{L}_p$  & 70.70$^*$$_{\pm 0.15}$ & 68.76$^{*}$$_{\pm 0.14}$ & 40.59$^{**}$$_{\pm 0.13}$ \\ 
w/o $\mathcal{L}_e + \mathcal{L}_p$    & 69.71$^{**}$$_{\pm 0.17}$ & 68.39$^{**}$$_{\pm 0.11}$ & 39.56$^{**}$$_{\pm 0.15}$ \\
w/o $_{\mathcal{D}_{domain}}$        & 70.91$^*$$_{\pm 0.13}$ & 68.62$^{*}$$_{\pm 0.19}$ & 40.54$^{*}$$_{\pm 0.19}$  \\
w/o $_{LoRA}$        & 70.30$^{**}$$_{\pm 0.11}$ & 64.80$^{**}$$_{\pm 0.12}$ & 40.05$^{**}$$_{\pm 0.21}$ \\
\bottomrule
\end{tabular}
}
\begin{tablenotes}
\item[a] \tiny{Results with standard deviation and significance testing between w/o* and LLama2 (*p$<$0.05, **p$<$0.01.)}
\end{tablenotes}
\label{tab3}
\end{table}

\begin{figure*}[]
    \centering
    \includegraphics[width=2.0\columnwidth]{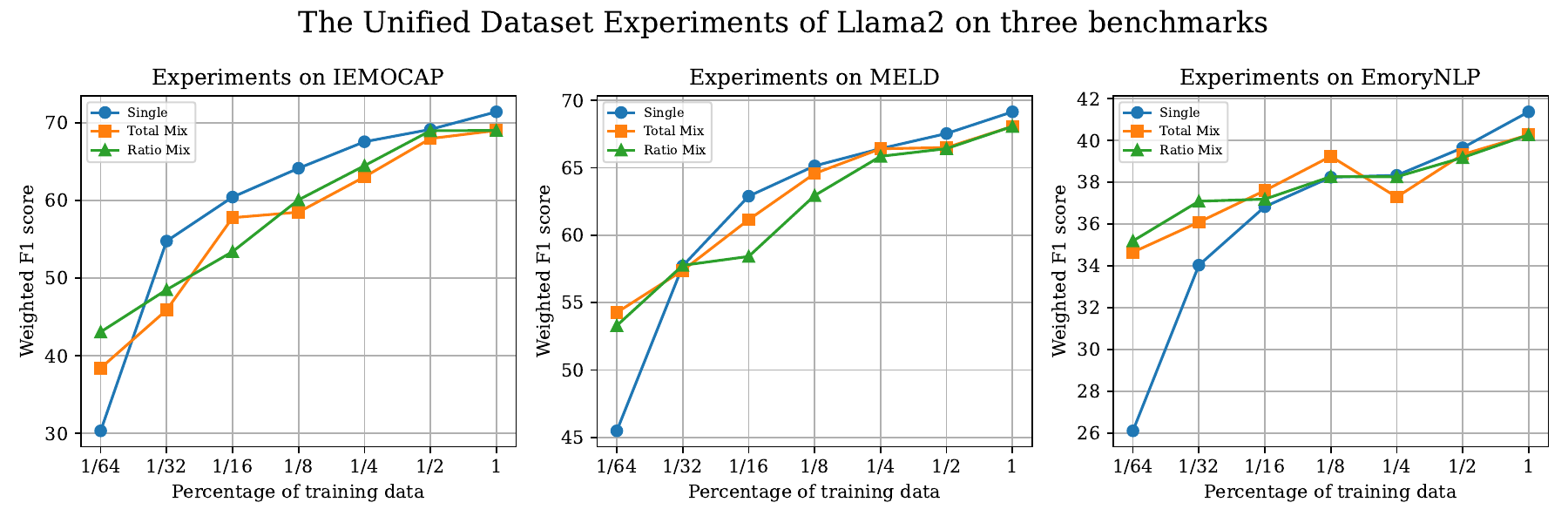}
    \caption{The data scaling analysis demonstrated on three benchmarks using different data mixing strategies}
    \label{fig:scaling2}
\end{figure*}

\section{Unified dataset Experiments}
In real-world scenarios, the ideal ERC model should be able to address ERC challenges across multiple domains, and even carry out open-domain ERC tasks. 
However, biases in data distribution and labeling make it challenging for small ERC models to achieve multi-domain capabilities,
To better simulate real-world scenarios, we first reconstruct three ERC datasets into a single ERC dataset (UIME) with unified labels based on the Emotion Wheel (Figure \ref{fig:Unified_mapping}), to better suit more industrial scenarios. 
\subsection{Unified Dataset Experiment Setup}
Within the settings of this experiment, all emotional labels across the datasets are standardized, and all speaker labels are also consolidated. The unification details of speaker labels and emotional labels can be refered to Appendix \ref{appendix:unifiedDetails}. 
Subsequently, we conduct data scaling experiments on the UIME.
To explore the impact of different sampling methods on the final performance, two data scaling approaches are experimented with: Total Mixing and Ratio Mixing.

In the ``Total Mixing'' approach, all subdatasets in UIME are first merged together, and then \{1, 1/2, 1/4, 1/8, 1/16, 1/32, 1/64\} amounts of data are randomly sampled separately from the merged data to fine-tune instructERC. Conversely, in the ``Ratio Mixing'' approach, \{1, 1/2, 1/4, 1/8, 1/16, 1/32, 1/64\} amounts of subdatasets are first randomly sampled separately, and then they are merged in accordance with their respective ratios to form the training data. Both approaches maintain the same quantity of the final training data.

The details of results are shown in Table \ref{tab6} in Appendix \ref{appendix:unifiedDetails}, and a more intuitive presentation is shown in Figure \ref{fig:scaling2}.

\subsection{The Robustness of InstructERC} 

As depicted in the Figure \ref{fig:scaling2}, 
Compared to the single dataset training setup, the performance of InstructERC, when fine-tuned on the UIME, has experienced a minor drop across three benchmarks. Specifically, there's a decrease of 2.4\% in IEMOCAP, 1.08\% in MELD, and 1.1\% in EmoryNLP.
However, a relatively high Weighted F1 score (W-F1) can still be maintained simultaneously on these three benchmarks, particularly the performance of MELD(68.07\%), which continues to surpass the SOTA level of all small models.
The results exhibits InstructERC 's exceptional robustness, which is capable of concurrently acquiring emotional paradigms from a multitude of distinct distributions
\footnote{The statistics of scaling analysis can be found in Table \ref{tab6}}.

\subsection{The Data Scaling Exploration}

The data scaling experiments are conducted on the unified dataset from 1 to 1/64.
As the scale of trainig data exponentially decreases from 1 to 1/32 within the range, the performance of the model on the three benchmarks exhibits a slight fluctuation in linear decline.


We are also surprised to discover that during the final stage of training data reduction from 1/32 to 1/64, the Total Mixing and Ratio Mixing strategies continue to exhibit a linear performance decline.
However, the performance of the model trained under the single method experiences a drastic drop, as depicted in Figure \ref{fig:scaling2}.
We posit that data from different scenarios endows the model with the capability to comprehend emotions from diverse perspectives.
This, in turn, allows the model to achieve robust enhancements under various data conditions.
Such mutual gain is particularly pronounced in low resource scenarios (1/64).
This is consistent with the findings of some existing explorations in large models \cite{dong2023abilities}.

\subsection{The Discussion of Mixing Strategies}
We have further investigated the impact of different mixing strategies on data scaling. 
The results displayed by different datasets on various mixing strategies can be interpreted from the following two perspectives:

\textbf{Data Representativeness:} In Total Mixing sampling, where each dataset's samples are equally likely to be selected, the unique traits of smaller datasets like IEMOCAP may be obscured by larger ones like MELD. In contrast, Ratio Mixinging sampling, which represents each dataset proportionally to its original sample size, may better highlight the characteristics and influence of smaller datasets.

\textbf{Effect of Class Imbalance:} In smaller datasets with internal class imbalances, Total Mixing sampling could exacerbate these imbalances. For instance, if IEMOCAP has a relatively smaller number of samples in a certain category, Total Mixing sampling might further intensify this imbalance during model training. Ratio Mixing sampling, however, better preserves the original class proportions of the datasets, potentially mitigating class imbalance impacts to a degree.

\section{Conclusion}
We introduce InstructERC, a novel approach that transforms the ERC task from a discriminative framework to a generative framework using LLMs. 
InstructERC presents a simple and effective retrieval template adapting to different conversation lengths.
Futhermore, we introduce two emotional alignment tasks 
to model speaker and complex conversation relationships.
InstructERC outperforms all previous models and achieve comprehensive SOTA results on three benchmarks. 
We also pioneer in unifying label mapping and modeling across these datasets, demonstrating the InstructERC's robust generalization capabilities. 
Our extensive analysis provides practical insights for implementing InstructERC in real-world ERC scenarios.
\section*{Limitation}
In this work, we focus solely on the textual aspects of these datasets. The exploration of multimodal aspects is reserved for future research.
We have conducted our explorations specifically on two representative large model frameworks, ChatGLM and LLaMA. Due to limitations in our graphics card capacity, the maximum parameter size of the large models we used does not exceed 7 billion.
    
\section*{Ethics Statement}
All the data sets we used for the experiment were published publicly. These data sets passed the ethical review at the time of publication. All the non-original methods and modules mentioned in this article have quoted other people's literature. All our science artifacts observe MIT licese.

\bibliography{custom}
\bibliographystyle{acl_natbib}

\clearpage
\appendix
\section{The Details of Unified Dataset Experiment Setup}
\label{appendix:unifiedDetails}


To further substantiate the efficacy and robustness of our framework,  
we conduct a compelling experiment involving a unified dataset.
Within the settings of this experiment, all emotional labels across the datasets are standardized, and all speaker labels are also consolidated.
Subsequently, we conduct data scaling experiments on the processed unified dataset.
The evaluation method employed in the experimental results, utilizing the weighted F1 score, aligned with the evalution method delineated in Section Experiments.

\begin{table}[]
\centering
\caption{Unified Label Mapping}
\scalebox{0.75}{
\begin{tabular}{ccccc}
\hline
Number                 & IEMOCAP            & MELD               & EmoryNLP                                & Final Emotion \\ \hline
\multicolumn{1}{c|}{1} & happy              & joyful             & \multicolumn{1}{l|}{joyful}             & joyful        \\
\multicolumn{1}{c|}{2} & sad                & sad                & \multicolumn{1}{l|}{sad}                & sad           \\
\multicolumn{1}{c|}{3} & neutral            & neutral            & \multicolumn{1}{l|}{neutral}            & neutral       \\
\multicolumn{1}{c|}{4} & angry              & angry              & \multicolumn{1}{l|}{mad}                & mad           \\
\multicolumn{1}{c|}{5} & excited            & N\textbackslash{}A & \multicolumn{1}{l|}{N\textbackslash{}A} & excited       \\
\multicolumn{1}{c|}{6} & N\textbackslash{}A & surprise           & \multicolumn{1}{l|}{powerful}           & powerful      \\
\multicolumn{1}{c|}{7} & scared             & fear               & \multicolumn{1}{l|}{frustrated}         & fear          \\   
\multicolumn{1}{c|}{8} & N\textbackslash{}A & N\textbackslash{}A & \multicolumn{1}{l|}{peaceful}           & peaceful      \\
\multicolumn{1}{c|}{9} & N\textbackslash{}A & disgust            & \multicolumn{1}{l|}{N\textbackslash{}A} & disgust       \\ \hline
\end{tabular}
}
\label{tab7}
\end{table}


We continue to use the previous datasets IEMOCAP, MELD, and EmoryNLP.
According to The Feeling Wheel \cite{willcox1982feeling} proposed in 1982, as shown in subfigure of Figure \ref{fig:Unified_mapping}, we align all emotional labels from three datasets with this standard, the details of which are shown in Tabel \ref{tab7}.
After completion of label mapping, there are a total of 9 types of emotional labels, which are \textit{ joyful, sad, neutral, mad, excited, powerful, fear, peaceful and disgust}. 
Furthermore, due to the uniqueness of character labels in each dataset, we have renumbered them using a One-hot encoding approach, as demonstrated in the "One-hot Speaker Label Mapping"  Table \ref{tab9}, which also is shown in subfigure of Figure \ref{fig:Unified_mapping}.
\begin{table}[]
\centering
\caption{One-hot Speaker Label Mapping}
\scalebox{0.9}{
\begin{tabular}{cccc}
\hline
Speaker label                 & IEMOCAP            & MELD               & EmoryNLP                                \\ \hline
\multicolumn{1}{c|}{1} & 1              & N\textbackslash{}A             & N\textbackslash{}A                     \\
\multicolumn{1}{c|}{\dots} & \dots      & N\textbackslash{}A                & N\textbackslash{}A                     \\
\multicolumn{1}{c|}{$n_1$} & $n_1$       & N\textbackslash{}A            & N\textbackslash{}A                   \\
\multicolumn{1}{c|}{$n_1+1$} & N\textbackslash{}A        & 1              & N\textbackslash{}A                    \\
\multicolumn{1}{c|}{\dots} & N\textbackslash{}A   & \dots &               {N\textbackslash{}A}          \\
\multicolumn{1}{c|}{$n_1+n_2$} & N\textbackslash{}A & $n_2$           & N\textbackslash{}A                 \\
\multicolumn{1}{c|}{$n_1+n_2+1$} & N\textbackslash{}A   & N\textbackslash{}A           & 1              \\   
\multicolumn{1}{c|}{\dots} & N\textbackslash{}A & N\textbackslash{}A     & \dots               \\
\multicolumn{1}{c|}{$n_1+n_2+n_3$} & N\textbackslash{}A & N\textbackslash{}A   & $n_3$           \\ \hline
\end{tabular}
}
\label{tab9}
\end{table}

We still utilize the LoRA method in PEFT to train InstructERC on the unified dataset, and the training results are evaluated on the three datasets respectively.
As mentioned above, these datasets have significant variations in sample size and class imbalance within each dataset. To explore the impact of different sampling methods on the final performance, two data scaling approaches were experimented with: Total Mixing and Ratio Mixing.

In the Total Mixing approach, all datasets are combined for uniform sampling. Conversely, in the Ratio Mixing approach, datasets are sampled separately and then combined. Both approaches maintain the same quantity of training data, but due to the larger absolute number of training samples in MELD and EmoryNLP, the Total Mixing approach results in a higher proportion of samples from these two datasets when varying data scaling is applied.

Total Mixing and ratio Mixing modes are applied proportionally across the entire training set, while still segregating a validation set and a test set. The reported results are obtained after training on a unified training set and then testing on separate test sets. The Single mode, on the other hand, involves training on individual training sets and then testing on their respective test sets.

Meanwhile, we design Total Mixing and Ratio Mixing experiments to explore the impact of different data mixing strategies and data quantities on the model.
On the basis of the following, we further explore the impact of data sampling ratio on the model's performance.The details of results are shown in Table \ref{tab6}, and a more intuitive presentation is shown in Figure \ref{fig:scaling2}.

\begin{table*}[htbp]
\caption{The Unified Dataset Experiments of Llama2 on three benchmarks}
\scalebox{0.73}{
\begin{tabular}{c|ccc|ccc|ccc}
\hline
\multirow{2}{*}{\textbf{Data Precent}} & \multicolumn{3}{c|}{\textbf{IEMOCAP  W-F1}} & \multicolumn{3}{c|}{\textbf{MELD W-F1}}      & \multicolumn{3}{c}{\textbf{EmoryNLP W-F1}}   \\ \cline{2-10} 
                                       & Total Mixing   & Ratio Mixing       & Single          & Total Mixing      & Ratio Mixing      & Single         & Total Mixing      & Ratio Mixing      & Single         \\ \hline
1                                      & 68.99       & 68.99           & \textbf{71.39}  & 68.07          & 68.07          & \textbf{69.15} & 40.27          & 40.27          & \textbf{41.37} \\
1/2                                    & 67.95       & 68.96           & \textbf{69.13}  & 66.50          & 66.42          & \textbf{67.54} & 39.18          & 39.33          & \textbf{39.65} \\
1/4                                    & 63.02       & 64.46           & \textbf{67.54}  & 66.41          & 65.85          & \textbf{66.42} & 38.26          & 37.29          & \textbf{38.33} \\
1/8                                    & 58.48       & 60.06           & \textbf{64.13}  & 64.57          & 62.94          & \textbf{65.14} & 38.27          & \textbf{39.24} & 38.24 \\
1/16                                   & 57.77       & 53.40           & \textbf{60.42}  & 61.15          & 58.42          & \textbf{62.89} & 37.19          & \textbf{37.60} & 36.83          \\
1/32                                   & 45.89       & 48.50           & \textbf{54.76}  & 57.38          & \textbf{57.76} & {57.72} & \textbf{37.09} & 36.09          & 34.03          \\
1/64                                   & 38.42       & \textbf{43.07}  & 30.34           & \textbf{54.26} & 53.29          & 45.48          & \textbf{35.19} & 34.65          & 26.10          \\ \hline
\end{tabular}
}
\label{tab6}
\end{table*}


\section{Related Works}
\label{appendix:related works}
\subsection{Emotion Recoginition in Conversation}
After more than a decade of development, the field of Emotion Recognition in Conversation (ERC) has seen many outstanding works. These can be broadly classified into three categories: Transformer-based, GNN-based, Recurrent-based.

Specifically, \textbf{Transformer-based} works \cite{li2020multi,song2022supervised,liu2023hierarchical,yingjian2023emotionic,chudasama2022m2fnet} attempt to establish long-range emotional correlations in conversational scenarios by directly adopting or modifying the original transformer block. These efforts have made significant contributions in this direction.

\textbf{GNN-based} works \cite{ghosal2019dialoguegcn,ishiwatari2020relation,shen2021directed,li2023graphcfc} extensively use graphs and edges to model interactions between people in conversational scenarios and the influences between different modalities. They employ various forms of multi-layer graph neural networks to fit potential conversational relations, effectively exploring this direction.

\textbf{Recurrent-based} works \cite{hu2023supervised,lei2023watch,majumder2019dialoguernn,hazarika2018icon,poria2017context} utilize various forms of RNNs, like LSTM and GRU, to model individual emotional states and global emotional impacts separately. They incorporate attention mechanisms or direct vector concatenation to represent personal and global emotional states collectively, marking effective exploration in this area.

\subsection{Large Language Models}
The emergence of large-scale language models (LLMs) have brought revolutionary transformation to the field of natural language processing (NLP) \cite{shen2023hugginggpt}. LLMs, such as GPT-3 \cite{brown2020language}, LLaMA \cite{touvron2023llama} and GPT-4 \cite{openai2023gpt4}, have demonstrated impressive abilities on various tasks, as well as the use of external techniques such as reinforcement learning from human feedback (RLHF) \cite{ouyang2022training}. LLMs based on generative framework even reformulate the multi modal perspective \cite{10.1145/3447548.3467206,zhang-etal-2023-pay,qiao2024we}. More recently, the NLP community has been exploring various application directions for LLMs. For instance, chain-of-thought prompting and RFT~\citep{wei2023chainofthought,yuan2023scaling,dong2024understand,li2024dotamath} enables LLMs to generate problem-solving processes step-by-step, significantly enhancing the model's reasoning ability. Researchers have utilized the interactive capabilities of LLMs to generate commands that invoke external tools for handling of downstream tasks~\citep{shen2023hugginggpt}. a series of works~\citep{le2022coderl,qiao2023making,dong2024self} utilize execution feedback from tools such as code executors to provide supervision for specific tasks.
Other researchers have proposed parameter-efficient fine-tuining (PEFT) to address the issue of excessive computational resource without sacrificing performance \cite{hu2021lora}.

\section{Datasets \& Baselines}

\subsection{Datasets}
\begin{table*}[h]
	\centering
  \caption{The statistics of datasets. $\rm avg\_utt$ denotes the average number of utterances in a conversation. } 
	\footnotesize
	\begin{tabular}{c|c|c|c|c|c|c|c|c|c|c}	
	\toprule
	\multirow{2}{*}{Datasets} & \multicolumn{3}{c|}{Conversations}  & \multicolumn{3}{c|}{Utterances} & \multirow{2}{*}{classes} & \multirow{2}{*}{type} & \multirow{2}{*}{avg\_utt} & \multirow{2}{*}{Evaluation} \\ 
		& Train & Val & Test & Train & Val & Test & & & & \\
		\midrule
		IEMOCAP     & 108   & 12  & 31  & 5163  & 647   & 1623 & 6 & two-person   & 47  & W-F1 \\ 	
		MELD        & 1038  & 114 & 280 & 9989  & 1109  & 2610 & 7 & multi-party  & 9   & W-F1 \\ 
		EmoryNLP    &  713  & 99  & 85  & 9934  & 1344  & 1328 & 7 & multi-party  & 11  & W-F1 \\ 
		\bottomrule
	\end{tabular}
  \label{tab:datasets_statics}	
\end{table*}

\label{appendix:datasets}
\textbf{IEMOCAP} \cite{busso2008iemocap} is a dataset recorded as dyadic conversational video clips with eight speaker participating in the training set while two speaker in testing set. 

\textbf{MELD} dataset \cite{poria2018meld} is a multimodal dataset that has been expanded from the EmotionLines dataset. MELD is obtained from the popular TV show \textit{Friends} and comprises over 1400 dialogues and 13000 utterances, each of which is labeled with emotion and sentiment classes. 

\textbf{EmoryNLP} \cite{zahiri2017emotion} is a dataset also collected from the TV series \textit{Friends}. The dataset comprises utterances that are categorized into seven distinct emotional classes.

This study exclusively focuses on the emotional classes and the text modality in these datasets. Moreover, we ensure consistency with COSMIC regarding the train/val/test splits.

\subsection{Baselines}
\label{appendix:baselines}
For discriminative ERC models, we selected several \textbf{SOTA} baseline for each method. For our reconstructed generative model, we chose four popular LLMs as backbones.

\textbf{Recurrent-based}: \textbf{1) EmotionIC} \cite{yingjian2023emotionic} uses IM-MHA and DialogGRU to capture contextual information in the dialogue, and SkipCRF to capture high-order dependencies between speakers for emotional flow simulation.
\textbf{2) SACL-LSTM} \cite{hu2023supervised} extracts structured representations using contrast-aware adversarial training and joint class-spread contrastive learning, an additional contextual adversarial training strategy to enhance context robustness. 

\textbf{Transformer-based}: \textbf{1) MPLP} \cite{lu-etal-2022-unified} is a framework that unifies multimodal sentiment analysis and emotion recognition in conversation tasks. This framework achieves this by performing modality fusion at both the syntactic and semantic levels, and by introducing contrastive learning between modalities and samples. \textbf{2) SPCL} \cite{song2022supervised} is a method that addresses imbalanced classification issues using Prototypical Network and contrastive learning, without the need for large batch sizes, and incorporates a difficulty measure function and curriculum learning to mitigate the effects of extreme samples.

\textbf{GNN-based}: \textbf{1) DualGATs} \cite{li2021past} uses a connected graph to enhance the targeted utterance with information from the past and future context, and utilizes CommonSense Knowledge (CSK) to enrich edges with knowledge representations.
\textbf{2) Skier} \cite{li2023graphcfc} is a module that efficiently models contextual and interactive information for ERC task. It uses multiple extractors and PairCC strategy to address the heterogeneity gap in multimodal fusion.

\textbf{LLM backbones:}
\textbf{1) ChatGLM-6B \& ChatGLM2-6B}: ChatGLM-6B is an open-source conversational language model \cite{du2022glm} for Chinese and English. It has 6.2 billion parameters and is optimized for Chinese QA. It has been trained on 1 trillion Chinese and English identifiers and further improved through various techniques. ChatGLM2-6B is the second generation of the model, pre-trained on 1.4 trillion Chinese and English identifiers with human preference alignment training. It extends the context window to 32K and speeds up inference with Multi-Query Attention.
\textbf{2) Llama-7B \&  Llama2-7B}: Llama-7B is the 7B parameters' version of the a collection of foundation language models \cite{touvron2023llama} ranging from 7B to 65B parameters, which is trained on trillions of tokens. Llama2-7B pretrained models are trained on 2 trillion tokens, and have double the context length than Llama 1. Its fine-tuned models have been trained on over 1 million human annotations.

\section{Implementation \& Discussion}
\subsection{Implementation Details}
\begin{table*}[htbp]
\small
\caption{The more detailed results and Statistics on three benchmarks.}
\centering
\scalebox{1.03}{
\begin{tabular}{c|c|c|c|c|c|c|c}
\toprule
 Dataset            & Parameters   & IEMOCAP         & MELD         & EmoryNLP  & Average   &  Extra      & Model type      \\
{Models}            &              & {W-avg F1}    & {W-avg F1}       & {W-avg F1}   &    {W-avg F1}         &  Knowledge  & \\
\midrule 
 \multicolumn{8}{c}{Small-scale Discriminant ERC-specific Model }       \\
\midrule 

{KET}$^*$           & 2.6M      & 59.56               & 58.18                      & 34.39              & 50.17                          & ConceptNet & transformer  \\      
{TODKAT}$^\dagger$  & 330M      & 61.33             & 65.47                      & 38.69              & 55.16                          & COMET & transformer  \\
{MTL}$^*$           & 1.2M      & -----             & 61.90                      & 35.92              & -----                          & \usym{2717} & transformer  \\
{CoG-BART}$^*$      & 415.1M    & 64.87             & 63.82                      & 37.33              & 55.34                          & \usym{2717} & transformer  \\
{M2FNet}$^*$        & -----     & 69.86             & 66.71                      & -----              & -----                          & \usym{2717} & transformer  \\
{SPCL}$^\dagger$    & 356.7M    & 68.42             & 66.13    & {\color{orange} \textit{40.25}}      & 58.26                  & \usym{2717} & transformer  \\
{Hidialog}$^*$      & -----     & -----             & {\color{orange}{66.96}}     & -----              & -----                          & \usym{2717} & transformer  \\
{SACL-LSTM}$^*$          & 2.6M      & 69.22             & 66.45                      & 39.65              & 58.44                          & \usym{2717} & recurrent    \\
{HCAN}$^\dagger$          & 3.5M      & 69.21              & 66.24                      & 39.67              & 58.37                          & \usym{2717} & recurrent    \\
{ICON}$^*$          & 0.5M      & 63.50            & -----                      & -----              & -----                          & \usym{2717} & recurrent  \\
{DialogueRNN}$^\dagger$   & 9.9M      & 64.65              & 65.30                      & 37.54              & 55.83                          & \usym{2717} & recurrent  \\
{DialogueCRN}$^\dagger$   & 3.3M      & 67.53              & 65.77                      & 38.79              & 57.36                          & \usym{2717} & recurrent  \\
{EmotionIC}$^*$     & -----         & 69.50             & 66.40          & 40.01  & {\color{orange} \textit{58.63}} & \usym{2717}  & recurrent    \\
{CauAIN}$^*$        & 6.1M      & 65.01                   & 64.89                      & 37.87              & 55.92                          & ATOMIC & recurrent  \\
{COIN}$^*$          & 0.5M      & 65.37                   & -----                      & -----              & -----                          & \usym{2717} & recurrent  \\
{COSMIC}$^\dagger$        & 11.9M     & 65.03 & 63.43                                        & 38.49              & 55.65                          & COMET & recurrent  \\
{DialogueGCN}$^\dagger$   & 2.1M      & 62.11                 & 62.68                      & 36.43              & 53.14                          & \usym{2717} & GNN  \\
{RGAT}$^*$          & 13M       & 65.22                     & 60.91                      & 34.42              & 53.52                          & \usym{2717} & GNN  \\
{SKAIG}$^*$         & -----     & 66.96                      & 65.18                      & 38.88              & 57.01                          & COMET & GNN  \\
{DAG-ERC}$^\dagger$ & 9.5M      & 66.54               & 63.36                      & 38.29              & 56.06                          & \usym{2717} & GNN  \\
{GraphCFC}$^*$      & -----     & 68.91              & 58.86                      & -----              & -----                          & \usym{2717} & GNN\\
\midrule
 \multicolumn{8}{c}{Small-scale Pretrained Language Model }       \\
\midrule
{KI-NET}$^*$        & 500M      & 67.00                     & 63.24                      & -----              & -----                          & ConceptNet & transformer \\
{DialogueXL}$^*$    & 510M      & 65.94                     & 62.41                      & 34.73              & 54.36                          & \usym{2717} & transformer\\
{EmoBERTa}$^*$      & 355M      & 68.57                       & 65.61                      & -----              & -----                          & \usym{2717} & transformer \\
{UniMSE}$^*$        & 220M      & {\color{orange} \textbf{\textit{ 70.66}}}  & 65.51      & -----              & -----                          & \usym{2717} & transformer   \\
\midrule
 \multicolumn{8}{c}{Zero-shot + InstructERC}       \\
\midrule
 ChatGLM $^\dagger$  & 12.5M(6B) & \textbf{\textit{38.6}}  & \textbf{\textit{38.8}}  & 19.6 & \textbf{\textit{32.33}} & \usym{2717} & LLM-based \\
ChatGLM2 $^\dagger$  & 12.5M(6B) & 21.1  & 21.8  & \textbf{\textit{24.4}} & 22.43  & \usym{2717} & LLM-based \\
Llama    $^\dagger$  & 12.5M(7B) & 0.753   & 9.12  & 5.31 & 5.06 & \usym{2717} & LLM-based\\
Llama2   $^\dagger$  & 12.5M(7B) & 2.774  & 16.28 & 8.36 & 9.46 & \usym{2717} & LLM-based\\
\midrule
 \multicolumn{8}{c}{LoRA + Backbone}       \\
\midrule 

ChatGLM $^\dagger$          & 12.5M(6B)     & 18.94   & {40.54} & {25.71}  & 28.07 & \usym{2717} & LLM-based \\
ChatGLM2$^\dagger$         & 12.5M(6B)     & 52.88    & {64.85}  & {37.69} & 51.80 & \usym{2717} & LLM-based \\
Llama$^\dagger$            & 12.5M(7B)     & 55.81    & \textbf{\textit{66.15}}  & {37.98} & 53.21 & \usym{2717} & LLM-based \\
Llama2$^\dagger$           & 12.5M(7B)     & \textbf{\textit{55.96}}  & {65.84} & \textbf{\textit{38.21}} & \textbf{\textit{53.33}} & \usym{2717} & LLM-based \\
\midrule
 \multicolumn{8}{c}{LoRA + InstructERC}       \\
\midrule 
 
ChatGLM$^\dagger$          & 12.5M(6B)     & 36.04                         & 46.41                 & 30.86             & 37.77  & \usym{2717} & LLM-based\\
ChatGLM2$^\dagger$         & 12.5M(6B)     & 67.54                          & 65.58              & 39.09             & 57.40  & \usym{2717} & LLM-based\\
Llama$^\dagger$            & 12.5M(7B)     & 64.17                          & 67.62              & 39.34             & 57.04  & \usym{2717} & LLM-based\\
Llama2$^\dagger$           & 12.5M(7B)     & {\color{red} \textbf{71.39}}    & {\color{red}\textbf{69.15}}   & {\color{red}\textbf{41.37}}    & {\color{red}\textbf{60.64}}  & \usym{2717} & LLM-based \\
\bottomrule
\end{tabular}
}
\begin{tablenotes}
    
\item[a] NOTE: The best-performing results of other models are highlighted in gold font, while SOTA results across all models are emphasized in red font. Models annotated with an * indicate results sourced from the model's paper, and a ($\dagger$) denotes results from reproductions conducted by the authors.
\end{tablenotes}
\label{detailed_results}
\end{table*}

\label{appendix:implementation}
We use ChatGLM and Llama as our backbone model. 
Considering the efficiency and effectiveness of Parameter-Efficient-Fine-Tuning (PEFT),
we adopt LoRA \cite{hu2021lora} and insert low-rank adapters after self-attention layers. We set the dimension of adapters to 16 a nd the learning rate to 2e-4. The learning rate is set to 2e-5 for all parameters' finetune. The histoical window is set to 1, 5, 12, 20 for iemocap, meld and EmoryNLP respectively for all experiments. The retrieval parameter “TopK” is set to Top1 emprically. 
The hypermeter $\alpha$ is set to 0.1 during training.
Greedy search is used during inference if not specified. Moreover, our experiments are conducted by taking the average of three runs with no hyperparameter searching. We train with FP16 precision on 4 $\times$ 80G Nvidia A100 GPUs. 


\subsection{Discussion with Discriminative ERC Models}
\label{appendix:Explanation_end_to_end}

\textbf{Problem definition.} 

In the discriminative framework, researchers first fine-tune an RoBERTA-style model with the context-free utterance, extract the feature vector at the CLS position as the input for the downstream ERC model. The aim is to map the feature vector of the given utterance to a scalar between 1 and $o$. 

In the generative framework based on LLMs, for a given utterance, we process it into formatted text according to the pre-designed template and input it into LLMs. The aim is to enable LLMs generate the most reasonable text emotional label, which must belong to the predefined text emotional label set $\mathcal{E} = \{e_1, e_2,..., e_o\}$. 

\textbf{Parameter Scales.} As shown in Table \ref{detailed_results}, we present the publicly available statistics for all trainable parameters across the models. Although the base architecture of our model is in the 6-7B parameter range, only 12.5M LoRA parameters are actively trained, which is feasible on a single GPU. For example, on the IEMOCAP dataset, our model typically converges by the 6th epoch, taking approximately 2 hours. The inference process requires about 10 minutes to handle 1000 samples. While our method is marginally slower than other approaches, such as the SPCL baseline which utilizes 356.7M training parameters, this speed reduction is not a significant drawback and remains manageable for most research contexts.

\textbf{Structural Complexity.} As shown in Table \ref{detailed_results} and Figure \ref{fig1}, taking the influential work such as COSMIC as an example, COSMIC fine-tuned RoBERTA on single-sentence dialogues, extracted its features, and encapsulated them into a dataset. Many works in the baseline are based on the feature dataset extracted from this work rather than the original text data for downstream model design. This means that these models (including but not limited to all the compared baselines which adopt this practice) need to use the single-sentence speech features fine-tuned with emotional labels during inference, which clearly does not conform to reality (the sentences that need to perform emotion recognition cannot access the gold emotional labels in advance). Furthermore, even if these single-sentence features do not need fine-tuning, it is still necessary to use Roberta to infer and obtain features.

In contrast, our InstructERC can directly input text and output emotional labels. Additionally, the InstructERC framework can be migrated to multiple datasets and combine datasets across multiple domains without modification, whereas discriminative models require manual changes to the architecture of the model, specifically the number of softmax classification neurons in the last layer, to perform multi-domain operations. In terms of scalability, the generative model InstructERC is clearly more practical than discriminative models.

\section{The Supplementary Experiments}
\subsection{The historical window exploration study}
\begin{table}[htbp]
\centering
\caption{The historical window exploration of Llama2 on three benchmarks.}
\scalebox{0.9}{
\begin{tabular}{cccc}
\toprule
 histoical          & IEMOCAP                    & MELD                       & EmoryNLP        \\ 
 window                  & {W-F1}       & {W-F1}       & {W-F1} \\ 
\midrule
 \multicolumn{4}{c}{LoRA + LLaMA2  + InstructERC}       \\
\midrule 
1             & 56.12$_{\pm 1.40}$ & {65.91}$_{\pm 0.46}$ & {38.32}$_{\pm 0.38}$  \\
5 & 68.65$_{\pm 0.32}$ & 66.97$_{\pm 0.21}$ & 40.48$_{\pm 0.23}$ \\
12 & \textbf{71.39}$_{\pm 0.10}$ & \textbf{69.15}$_{\pm 0.09}$ & \textbf{41.37}$_{\pm 0.11}$  \\
20 & 71.01$_{\pm 0.12}$ & 68.75$_{\pm 0.12}$ & 40.56$_{\pm 0.15}$ \\
\bottomrule
\end{tabular}
}
    
\vspace{-0.2cm}
\label{tab8}
\end{table}

\label{appendix:historical}
In the historical window exploration shown as Table \ref{tab8}, we examine how different sizes of historical windows affect emotion recognition tasks. Due to token limitations, we set the upper limit for conversational turns to 20. This is an upgrade from earlier, smaller Pretrained Language Models (PLMs, e.g. Roberta\cite{liu2019roberta}), which only support up to 5 turns. We find that a window of 12 turns is optimal for capturing the necessary historical context. In general, expanding the count of historical turns aids in enhancing the accuracy of emotion detection, a trend that is readily observable in the IEMOCAP dataset featured long-term turns. However, there's a point where adding more historical turns doesn't lead to better results and might even harm performance, especially for datasets like MELD and EmoryNLP, which have an average length of 6 to 7 turns. However, these insights are beyond the reach of smaller PLMs that top out at 5 turns.

\subsection{The Exploration Experiments on $\alpha$}

\begin{table}[htbp]
\centering
\caption{The exploration experiments on $\alpha$.}
\scalebox{1.0}{
\begin{tabular}{cccc}
\toprule
 $\alpha$          & IEMOCAP                    & MELD                       & EmoryNLP        \\ 
                   & {W-F1}       & {W-F1}       & {W-F1} \\ 
\midrule
 \multicolumn{4}{c}{LoRA + LLaMA2  + InstructERC}       \\
\midrule 
0 & 70.50 & 68.97 & 40.78 \\
0.05 & 70.67 & 69.03 & 40.91 \\
0.1 & 71.39 & 69.15 & 41.37 \\
0.2 & 71.14 & 68.54 & 40.63 \\

\bottomrule
\end{tabular}
}
    
\vspace{-0.2cm}
\label{tab10}
\end{table}

Shown as Table \ref{tab10}, the influence of alpha on InstructERC's performance varies across different datasets due to their unique characteristics. In general, as alpha increases, its contribution to model performance also increases, peaking at alpha=0.1. Specifically, in the IEMOCAP dataset, characterized by longer dialogues averaging 47 turns, even when alpha exceeds 0.1 significantly, there is no significant decrease in performance. However, in datasets like MELD and EmoryNLP, which have shorter dialogues averaging 7 turns, an alpha value of 0.2 can lead to a negative impact, particularly evident in MELD. Therefore, careful consideration is necessary when selecting alpha values for different datasets.

This phenomenon can be explained as follows: In the IEMOCAP dataset, with its longer dialogues, emotional changes occur relatively slowly. In contrast, datasets like MELD and EmoryNLP, sampled from the sitcom ``Friends'', feature many brief and intense emotional shifts. Excessive reduction in the weight of emotion impact prediction may cause the model to overly emphasize the influence of past utterances on current emotion judgment, which may not be suitable for MELD and EmoryNLP.

\subsection{Label Ablation Experiments}
\label{appendix:label_ablation}
To further explore the impact of using the same or unrestricted emotional labels at different stages during the demonstration retrieval process on final performance, we designed experiments as shown in the table \ref{tab:dataset_performance}, where \texttimes  represents not using the same labels, and \checkmark represents using the same labels. Our conclusions are as follows:

\textbf{Impact of Label Restrictions:} The performance consistently improves across all datasets when moving from unrestricted to restricted labels in both training and inference. This suggests that restricting labels helps the model learn more robust features that are better at generalizing during inference.

\textbf{Comparison Across Datasets:}
IEMOCAP: Shows a steady increase in W-F1 scores as restrictions are applied first in training and then in both training and inference. The improvement from fully unrestricted to fully restricted is 1.54 points.
MELD: Similar to IEMOCAP, restricted training and inference show a noticeable improvement. The gain from the least to the most restricted setup is 2.54 points, indicating a potentially more significant impact of label restriction in emotionally complex interactions, possibly due to MELD's diverse emotional content and real-life scenarios.
EmoryNLP: This dataset shows the lowest overall scores but follows the same trend. The increase is 2.14 points from no restrictions to full restrictions. Given the smaller base score, this improvement is quite significant, emphasizing how crucial precise label handling is in models trained on this data.

\textbf{Fairness and Performance Trade-offs:}
The best results obtained by using restricted labels in both phases might not be fair or realistic for real-world applications, where the model shouldn’t have prior knowledge of the emotional context. This indicates a need for models that perform well under unrestricted conditions.
The performance drop when moving to unrestricted labels in inference underscores the challenge in generalizing the learned emotional cues without specific hints, highlighting a potential area for further research in enhancing model robustness.

\begin{table}[]
\centering
\caption{Dataset performance with various restrictions on labels during training and inference.}
\scalebox{0.9}{
\begin{tabular}{c|c|c|c}
\toprule
\textbf{Dataset} & \textbf{Training} & \textbf{Inference} & \textbf{W-F1} \\ \midrule
IEMOCAP & \texttimes & \texttimes & 70.71 \\
IEMOCAP & \checkmark & \texttimes & 71.39 \\ 
IEMOCAP & \checkmark & \checkmark & 72.25 \\ \hline
MELD & \texttimes & \texttimes & 68.52 \\ 
MELD & \checkmark & \texttimes & 69.15 \\ 
MELD & \checkmark & \checkmark & 71.06 \\ \hline
EmoryNLP & \texttimes & \texttimes & 40.54 \\ 
EmoryNLP & \checkmark & \texttimes & 41.37 \\
EmoryNLP & \checkmark & \checkmark & 42.68 \\ \bottomrule
\end{tabular}
}
\label{tab:dataset_performance}
\end{table}

\subsection{All Parameters vs Parameter Efficiency }
\label{appendix:parameters}




\begin{figure*}[t]
\centering
\includegraphics[width=2.0\columnwidth]{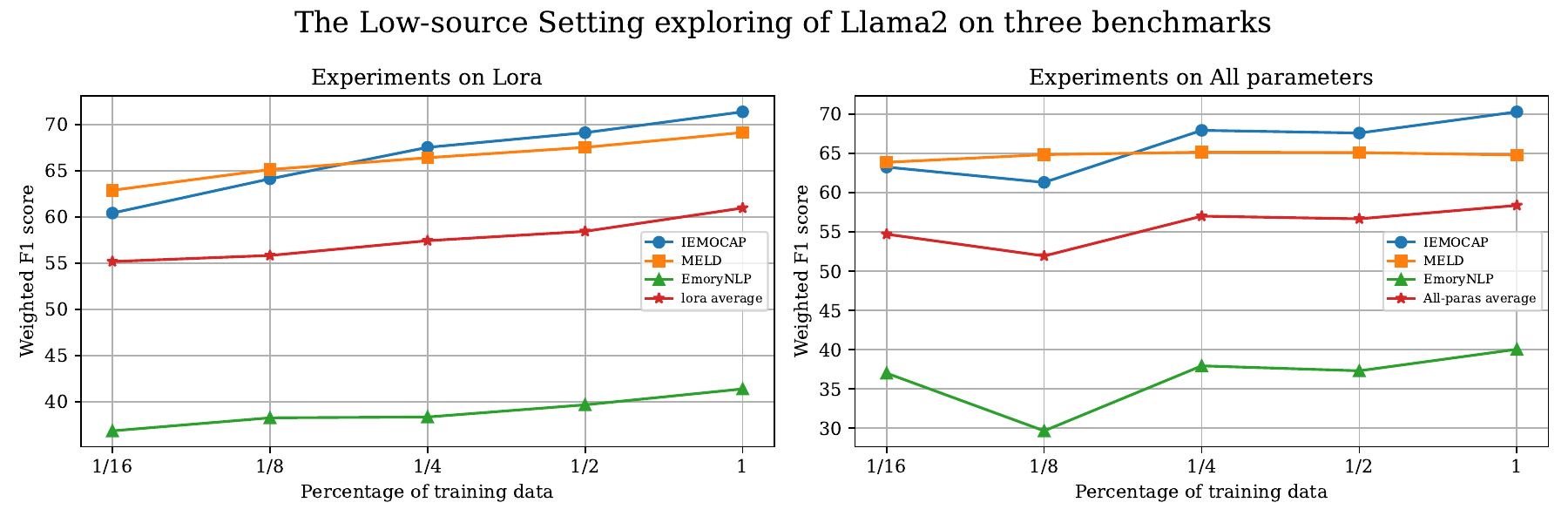} 
\caption{The scaling of data and performance for different parameter fine-tuning settings (LoRA \& All Parameters)}
\label{fig4}
\end{figure*}

In order to investigate the effect of different parameter fine-tuning methods on the ERC task, we conducted comparative experiments in Table \ref{tab4}. We have the following observations:

(1) The all parameter fine-tuning performs weaker than LoRA's fine-tuning on all backbones on average performance (especially ChatGLM with a 9.32 \% improvement). It is worth noting that the best performance of the full parameter method is often achieved in the first 1-3 epochs in the experiment. These findings demonstrate that parameter-efficient methods are more suitable for LLMs in ERC tasks.
\begin{table}[]
\caption{The comparison results of different parameter fine-tuning settings on three benchmarks.}
\scalebox{0.75}{
\begin{tabular}{ccccc}
\toprule
 Dataset          & IEMOCAP                    & MELD                       & EmoryNLP                   & Average \\
{Models}                   & {W-F1}       & {W-F1}       & {W-F1}       & {W-F1} \\
\midrule
 \multicolumn{5}{c}{All parameters + InstructERC}       \\
\midrule 

ChatGLM$^\dagger$                       & {33.94} & {37.96} & {13.25} & 28.38 \\
ChatGLM2$^\dagger$                      & {70.05} & {63.24}  &\textbf{\textit{38.77}} & 57.35 \\
Llama$^\dagger$                         & \textbf{\textit{69.38}} & \textbf{\textit{66.01}}  & {40.21} & \textbf{\textit{58.53}} \\
Llama2$^\dagger$                        & {70.30} & {64.80} & {40.05} & 58.38 \\
\midrule
 \multicolumn{5}{c}{LoRA + InstructERC}       \\
\midrule 
 
ChatGLM$^\dagger$                               & 36.04                   & 46.41                 & 30.86             & 37.77 \\
ChatGLM2$^\dagger$                               & 67.54                     & 65.58              & 39.09             & 57.40 \\
Llama$^\dagger$                                  & 69.71                     & 68.89              & 39.90             & 59.50 \\
Llama2$^\dagger$                                 & \textbf{71.39}         & \textbf{69.15}        & \textbf{41.37}    & \textbf{60.64} \\
\bottomrule
\end{tabular}
}
\label{tab4}
\vspace{-0.2cm}
\end{table}

(2) From the perspective of model structure, the average performance of full parameter ChatGLM even decreases compared to the zero-shot results in Table \ref{tab2} (from 32.33\% to 28.38\%), while replacing it with LoRA brings a significant improvement (from 32.33\% to 37.77\%). Other decoder-only backbones do not show such drastic performance fluctuations, which further indicates that the prefix-decoder paradigm is unstable in ERC tasks compared to the casual decoder, and parameter-efficient frameworks can effectively alleviate this problem.

(3) From the perspective of datasets, compared to full parameter fine-tuning, the performance gain of the LoRA method in MELD and EmoryNLP is significantly greater than that in IEMOCAP. We believe that this is related to the characteristics of thees datasets: IEMOCAP has long dialogue texts and multiple conversation rounds, these strong supervision signals lead to good performance in both settings. However, MELD and Emory have fewer dialogue rounds, diverse speakers, and imbalanced categories. Low-parameter methods can effectively prevent LLMs from overfitting to certain semantic patterns of dialogues format and speaker's habits, thereby enhancing the generalization ability of emotion recognition in conversation.

\subsection{Scaling Analysis in Low-source Scenario}
\label{appendix:single-scaling}




In this section, we gain an insight into the scaling analysis of data and performance for different parameter fine-tuning settings (LoRA \& All Parameter), as shown in Figure \ref{fig4}.

\textbf{Parameter-efficient Scaling Analysis}: On the IEMOCAP dataset, our scaling curve initially increases (from 1/16 to 1/4) and then stabilizes. This may be because the dataset has long dialogue texts and multiple dialogue rounds, leading to increased diversity with the addition of early data. However, as the supervision signal strengthens, the performance gain gradually weakens. For datasets with fewer dialogue rounds and imbalanced categories, such as MELD and EmoryNLP, our method only yields a small gain in extremely low-resource scenarios (from 1/16 to 1/4) and achieves a relatively stable performance improvement with the increase of data (from 1/2 to 1). This finding supports the idea that when a unit-scaling of data only provides weak supervision signals, the data size needs to exceed a certain threshold (1/4 - 1/2) to achieve significant improvement. 

\textbf{Full-Parameter Scaling Analysis}: 
The scaling curves of full-parameter settings on the IEMOCAP and EmoryNLP datasets showed significant fluctuations and performance degradation in two intervals (from 1/16 to 1/8, 1/4 to 1/2) compared to LoRA. Fine-tuning large models with all parameters may cause redundant parameters to overfit the patterns in the current dialogue, which hinders the model's ability to generalize new supervised signals as data volume increases. The MELD dataset also exhibited performance degradation with data augmentation (from 1/4 to 1). These findings demonstrate the stability and robustness of parameter-efficient fine-tuning in the ERC task, providing empirical guidance for large models in industrial interfaces with ERC tasks of varying data characteristics.

\end{document}